\documentclass[10pt, a4paper]{article}

\usepackage[]{lrec-coling2024} % this is the new style

\usepackage{hyperref}

\usepackage{multirow}
\usepackage{multicol}
\usepackage{color,soul}
\setul{0.5ex}{0.3ex}
\setulcolor{red}
\usepackage{subcaption}
\usepackage{cleveref}
\usepackage{amssymb}% http://ctan.org/pkg/amssymb
\usepackage{pifont}% http://ctan.org/pkg/pifont
\usepackage{algorithm} 
\usepackage{algpseudocode} 
\usepackage{caption}
\usepackage{bbm}
\usepackage{todonotes}
\usepackage{float}
\definecolor{taa-color}{rgb}{0,0,1}
\definecolor{kmy-color}{rgb}{0.858, 0.188, 0.478}

\usepackage{cleveref}
\definecolor{darkblue}{rgb}{0, 0, 0.5}
\hypersetup{colorlinks=true, citecolor=darkblue, linkcolor=darkblue, urlcolor=darkblue}

\title{Granular Change Accuracy: A More Accurate Performance Metric for Dialogue State Tracking}

\name{Taha Aksu\textsuperscript{$\dagger\ddag * $,}, Nancy F. Chen\textsuperscript{$\ddag$}}

\address{
         \textit{$\dagger$} National University of Singapore (NUS), Singapore\\
          \textit{$\ddag$} Institute for Infocomm Research ($I^2R$), A*STAR, Singapore \\
          \texttt{*taksu@u.nus.edu}}

\abstract{
Current metrics for evaluating Dialogue State Tracking (DST) systems exhibit three primary limitations. They: $i$) erroneously presume a uniform distribution of slots throughout the dialog, $ii$) neglect to assign partial scores for individual turns, $iii$) frequently overestimate or underestimate performance by repeatedly counting the models' successful or failed predictions.
To address these shortcomings, we introduce a novel metric: Granular Change Accuracy (GCA). GCA focuses on evaluating the predicted changes in dialogue state over the entire dialogue history. Benchmarking reveals that GCA effectively reduces biases arising from distribution uniformity and the positioning of errors across turns, resulting in a more precise evaluation. Notably, we find that these biases are particularly pronounced when evaluating few-shot or zero-shot trained models, becoming even more evident as the model's error rate increases. Hence, GCA offers significant promise, particularly for assessing models trained with limited resources. Our GCA implementation is a useful addition to the pool of DST metrics.
 \\ \newline }

\begin{document}

\maketitleabstract

\section{Introduction}
% Statement introducing broad research area
Dialogue State Tracking (DST) is the task of extracting user preferences from a Task-Oriented Dialogue (TOD) to accomplish a task such as booking a hotel room~\cite{henderson-etal-2014-second}. The community has adopted a several different metrics to evaluate model performances on this task \cite{ye-etal-2022-assist,feng-etal-2022-dynamic, zhu-etal-2022-continual,hung-etal-2022-ds}; however, these metrics employ some weaknesses that can result in an imbalanced assessment, such that strong systems receive poor scores and vice versa.

\begin{figure}[hbt!]
\centering
\includegraphics[width=0.4\textwidth]{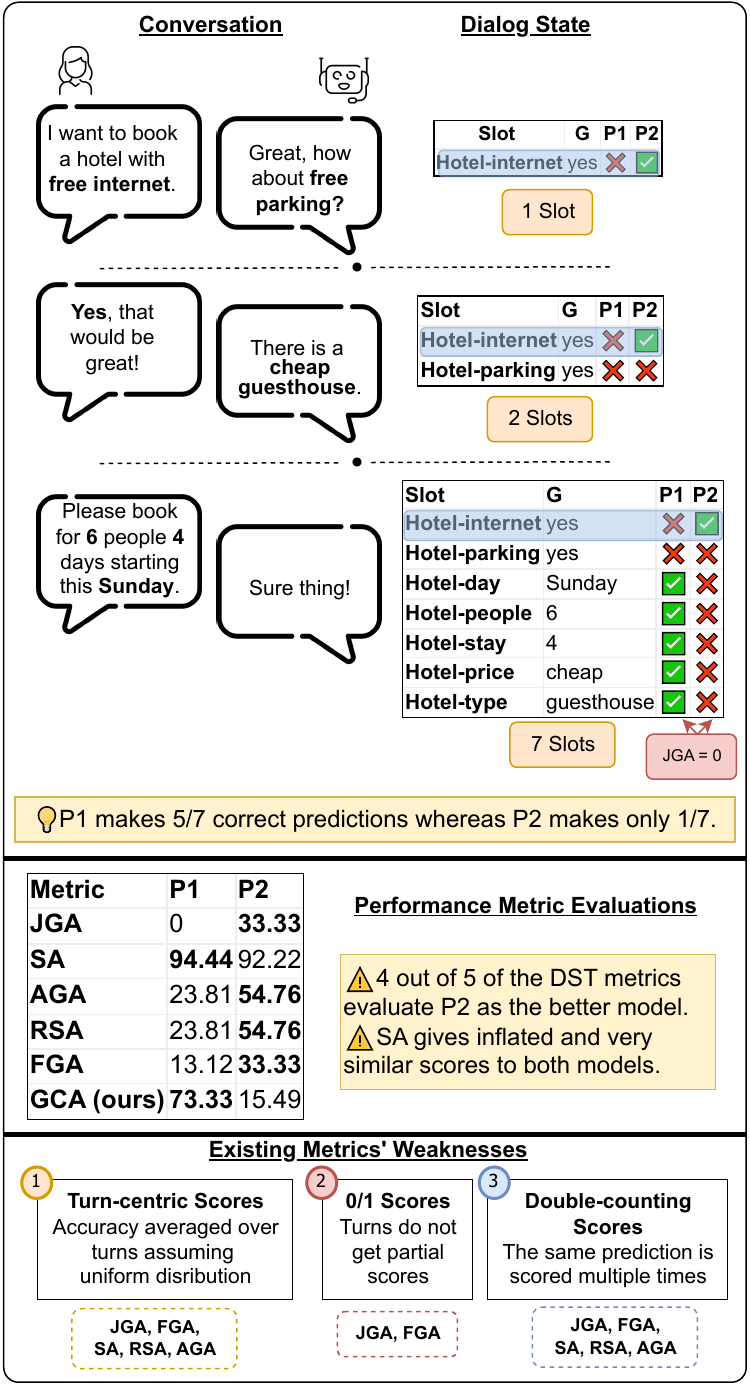}
\caption{Sample task-oriented dialogue with ground truth belief state $G$, and two belief state predictions $P1$ and $P2$.}
\label{Fig:main_fig}
\end{figure}
% Min2: BUG you use $P^1$ here in text but $P1$ in the Fig and in the caption.  Make consistent.
% Taha: Corrected
\Cref{Fig:main_fig} presents a sample TOD with two sets of DST predictions, $P1$ and $P2$. $P1$ predicts five of seven slots correctly whereas $P2$ only predicts one correctly. However, the metrics of Joint Goal Accuracy (JGA;~\citealt{henderson-etal-2014-second}), Flexible Goal Accuracy (FGA;~\citealt{dey-etal-2022-towards}), Relative Slot Accuracy (RSA;~\citealt{kim-etal-2022-mismatch} ) and Average Goal Accuracy (AGA;~\citealt{SGD}) evaluate the latter $P2$ as the better prediction. Just as problematic, Slot Accuracy (SA;~\citealt{wu-etal-2021-transferable}) gives inflated and similar scores to both predictions. The bottom of the figure depicts the source of these miss-evaluations in the form of weaknesses that these metrics employ.
% Min2: this is now much crisper and clear about the problems and your proposed solution.  Well done.
% Taha: Thanks Prof.!

% Min: Use back the same words or phrase in Fig 1. 
% Min: Consider moving #3 sentence to start a new paragraph
Firstly, their scores are turn-centric, \textit{i.e.} they uniformly treat each turn, averaging their accuracies, even when certain turns involve more slots and are inherently more challenging to predict. Secondly, their turn evaluation is limited to 0/1 scoring lacking a mechanism to assign partial credit for turns that have only a subset of slots predicted correctly. Lastly, they double-count scores, \textit{i.e.} they have a tendency to repeatedly penalize or reward the same predictions made in earlier turns.

% Recognizing the assumptions of DST task benchmarks is pivotal in understanding the final weakness. All DST datasets ( \textit{e.g.} MultiWOZ and SGD) operate under the assumption of a flawless DST model guiding the conversation. Each subsequent turn is based on a perfect model's output, so errors by the model aren't corrected or hinted at in later turns. In real-world scenarios, repeated mistakes might be more penalized due to their impact on user experience. However, these datasets don't account for compounded errors as in live situations. Metrics such as JGA and FGA, aiming to mirror real-world implications, sometimes introduce biases by over-penalizing early errors. This can skew model evaluations, making it essential to understand this distinction when using these benchmarks.

Such shortcomings make the current metrics sensitive to two spurious characteristics of model predictions: (1) the timing of mistakes (whether early or late in the dialog) and (2) the spread of errors among turns. To address these, we present Granular Change Accuracy (GCA). GCA gauges the variance between predictions and the actual belief states each turn, prevents repeated counting of predictions, and ensures precise evaluation by averaging over state alterations rather than turns.

% A summary of how the method was evaluated and the outcomes.
% Min2: You'll want to submit the code as supplemental material.  Maybe even hint this in the abstract ``Our available implementation is better in evaluating ...''
We evaluate GCA on the MultiWOZ~2.1 and SGD datasets
\cite{eric-etal-2020-multiwoz,rastogi2020towards}
% \citeplanguageresource{Multiwoz2_1}
, conducting benchmarking experiments with popular baselines and show that GCA positions in the middle of the spectrum, more optimistic than JGA and FGA's strict penalizing scheme, but not as inflated as SA and AGA. 
% We further conduct a qualitative analysis proving that GCA is 0.1 less correlated by the position of mistakes in the dialogue and 0.29 less correlated by the distribution uniformity of mistakes compared against the recent FGA metric with a  difference.

% We further conduct a qualitative analysis proving that GCA is significantly less correlated by the position of mistakes in the dialogue and by the distribution uniformity of mistakes compared against the most recent FGA metric. We also conduct experiments on few and zero-shot scenarios and find that the difference between GCA and the majority of existing metrics proved to be larger as training resources decreased. As models trained with fewer data tend to make more mistakes, they are affected more by these weaknesses. 
Our contributions are four-fold:
\begin{itemize}
    \item \textbf{Detailed Analysis of current metrics:} We perform a thorough examination of existing metrics, depicting their inherent biases sourced from three weaknesses depicted above.
    \item \textbf{Granular Approach:} We introduce Granular Change Accuracy (GCA)\footnote{The code is available at \url{https://github.com/cuthalionn/Granular_Change_Accuracy}}, a new DST evaluation metric that focuses on capturing belief state changes rather than a simple turn-by-turn assessment, effectively addressing the weaknesses prevalent in traditional metrics.
    \item \textbf{Comprehensive Benchmarking:} We evaluate GCA against other DST metrics on MultiWOZ 2.1 and SGD datasets, showcasing its superior balance in terms of evaluation accuracy. We further prove that GCA is significantly less correlated by the position of mistakes in the dialogue and by the distribution uniformity of mistakes compared to the most recent FGA metric.
    \item \textbf{Few-shot \& Zero-shot Experiments:} We shine a light on the heightened discrepancies between GCA and traditional metrics in low-resource settings, emphasizing the increased effect of the identified weaknesses when models are trained with less data.
\end{itemize}
% Thus we believe GCA  will be more accurate in evaluating few- and zero-shot models.

\section{Task Definition}

% DST is the task of extracting/generating the slot values for predefined slot labels specific to each domain, such as \textit{restaurant-food: Indian} in the restaurant domain. We refer to a slot label/value as simply \textit{slot} and \textit{value} in this paper. A task-oriented dialogue is represented as $D = \{(S_0,U_0,BS_0), ... , (S_{n-1},U_{n-1},BS_{n-1})\}$ where $S_i$ and $U_i$ form the $i_{th}$ turn pair and are system and user utterances, respectively; $BS_i$ is the belief state of the $i_{th}$ turn pair; and $n$ is the number of turn pairs. Each turn pair can incorporate zero or more slot--value pairs, and these are summarized in the dialogue state, i.e. $BS = \{(S_0:V_0), ... , (S_m:V_m)\}$ where $(S_j:V_j)$ is the $j_{th}$ active slot--value pair and $m$ is the number of slots predicted to have an active value in the current turn. Thus $m \leq K$ where $K$ is the number of pre-defined slots in the dataset (\textit{e.g.} 30 for MultiWOZ). The remaining $K-m$ slots are assigned a ``none'' value, indicating their inactivity in the current turn. It is important to note that the dialogue state accumulates throughout the conversation, meaning that predictions made in previous turns persist unless a new value is predicted, including ``none'' values.
DST involves extracting/generating slot values for specific slot labels in each domain (e.g., \textit{restaurant-food: Indian}). A task-oriented dialogue is represented as triplets of system and user turn pairs, and the turn belief state denoted as
% $D = {(S_0,U_0,BS_0), ... , (S_{n-1},U_{n-1},BS_{n-1})}$,
\begin{equation}
D = \left\{ (S_0, U_0, BS_0), \ldots , (S_{n-1}, U_{n-1}, BS_{n-1}) \right\}
\end{equation}
, where $S_i$ and $U_i$ are system and user utterances in the $i_{th}$ turn, and $BS_i$ is the belief state. Each turn pair can have zero or more slot--value pairs, summarized as 
% $BS = {(S_0:V_0), ... , (S_m:V_m)}$
\begin{equation}
BS = \left\{ (S_0:V_0), \ldots , (S_m:V_m) \right\}
\end{equation}
, where $(S_j:V_j)$ represents an active slot--value pair (\textit{i.e.} all pairs where the value is not ``none'') and $m$ is the number of active slots in the current turn. 
The remaining inactive slots are assigned a ``none'' value. Previous turn predictions persist unless new values are predicted, including ``none'' values.

\section{Related Work}

The evaluation of Dialogue State Tracking (DST) systems has seen various metrics proposed over the years, aiming to capture the nuances and complexities of dialogs. Among these, two metrics stand out due to their widespread adoption and historical significance: Joint Goal Accuracy (JGA) and Slot Accuracy (SA).
% The two most commonly reported DST metrics are JGA and SA.
\paragraph{Joint Goal Accuracy~\cite{henderson-etal-2014-second}}
computes the ratio of turn--pair slots that are correctly predicted across all turn pairs. For a prediction to be deemed correct, all slot--value sets in a turn--pair must align between the predicted and the ground truth belief states. The metric is formulated as:
% $JGA = \frac{\sum_{t=0}^{n}{(\mathbbm{1} \; | \; G_t=P_t)}}{n}$
\begin{equation}
JGA = \frac{\sum_{t=0}^{n}{(\mathbbm{1} \; | \; G_t=P_t)}}{n}
\end{equation}
, where $G_t$ and $P_t$ represent the ground truth and predicted belief states, respectively. One shortcoming of JGA is its tendency to underestimate results by not affording partial credit to turns. This shortcoming can be observed for the example in \Cref{Fig:main_fig}, JGA scores $P2$ higher despite $P1$ obviously performing better.

\paragraph{Slot Accuracy~\cite{wu-etal-2021-transferable}}
quantifies accuracy across all pre-defined slot labels, incorporating even those slots with a "none" value. It is expressed as:
% M denotes
% the number of missed slots that the model does not
% accurately predict among the slots included in the
% gold state, and W denotes the number of wrongly
% predicted slots among the slots that do not exist in
% the gold state.

% $SA = \frac{\sum_{t=0}^{n}{\frac{K - M_t - W_t}{K}}}{n}$
\begin{equation}
SA = \frac{\sum_{t=0}^{n}{\frac{K - M_t - W_t}{K}}}{n}
\end{equation}
where $M_t$ and $W_t$, are the number of missed predictions and wrong predictions (including slots that do not exist in the ground truth belief state) in turn $t$, and $K$ is the total number of slots specified in the dataset (for instance $K=30$ for MultiWOZ 2.1 dataset). A key limitation of SA is its inclination to overestimate model performance, largely because it rewards models for correctly identifying inactive slots. For the dialogue in \Cref{Fig:main_fig} for instance slot accuracy scores for $P1$ and $P2$ are very close and inflated despite the large difference they show in actual performance.
% Because SA considers all predefined slots at every turn it tends to overestimate the performance of models.

% JGA underestimates results since it denies partial credit from turns; whereas SA overestimates, as it rewards models for inactive slots.
Given the identified weaknesses in JGA and SA, subsequent research efforts have introduced alternative metrics such as Average Goal Accuracy, Relative Slot Accuracy, and Flexible Goal Accuracy to better capture DST performance nuances.

\paragraph{Average Goal Accuracy~\cite{SGD}}
differs from earlier metrics because it evaluates only the performance of turns with active slots; i.e., if a turn does not have any ground truth values, it will be discarded during the evaluation. It calculates a recall value for all turns with non-empty ground truth belief states and returns the average. AGA takes an average over each turn, leading to the recurring inclusion of each turn's mistakes or accurate predictions. AGA results for $P1$ and $P2$ at \Cref{Fig:main_fig} show that AGA also mistakenly chooses $P2$ as the better performing model. 
% \begin{equation}
%     AGA = \frac{\sum_{t=0}^{n}{(\frac{G_t \cap P_t}{|G_t|} \; | \; |G_t| \geq 1)}}{\sum_{t=0}^n{(\mathbbm{1} \; | \;  |G_t| \geq 1)}}
% \end{equation}

\paragraph{Relative Slot Accuracy~\cite{{kim-etal-2022-mismatch}}} addresses the overestimation problem in SA by calculating the score over active slots in the turn. 
% $RSA = \frac{{T^{*} - M - W}}{T*}$, 
\begin{equation}
RSA = \frac{{T^{*} - M - W}}{T*}
\end{equation}
, where $RSA= 0$ if $T^{*}$ = 0,
% and $T^{*}$ is the number of unique slots in the predicted and ground truth turns, $M$ is the number of missed slots, and $W$ is the number of wrong predictions.
and $T^{*}$, $M$ and $W$ are the number of unique slots in the predicted and ground truth turns, missed slots, and wrong predictions (including slots that do not exist in the gold belief state), respectively.  Despite its advancement over SA, RSA still computes the average over each turn, resulting in the repeated counting of errors and accurate predictions for each turn. $P1$ in \Cref{Fig:main_fig} gets a lower RSA score compared to $P2$. Thus it is evident that the RSA metric inherits similar weaknesses as AGA.

\paragraph{Flexible Goal Accuracy~\cite{dey-etal-2022-towards}}
adapts JGA by treating mistakes in the current turn as zero, while mimicking JGA's behavior. Yet, if all current turn slot values are correctly predicted with a mistake 
% Min2: ``propagated''?
% Taha: Sure
propagated from a previous turn, FGA applies a penalty that decreases over time with the decay ratio parameter $\lambda$. Although FGA softens the strict evaluation of JGA, it inherits a significant limitation; even a single error in the current turn results in the metric disregarding all other correct predictions. It is also evident from the FGA results in \Cref{Fig:main_fig}, where $P2$ again gets the higher score. Note that FGA's more flexible definition can be observed in the increased performance of $P1$, however, it still evaluates $P2$ to be the outperforming model out of the two. \\

% Closing a gap.
% Reverse the argument to make it easier to read -> We need a new metric because ...
% 2) RSA and AGA's reliance on turn-based averaging, combined with FGA's vulnerability to individual mistakes, 1) highlights the necessity for a metric that 3) comprehensively addresses dialogue dynamics without excessively penalizing or rewarding models.
The preceding discussion underscores the need for a fresh metric. Both RSA and AGA predominantly rely on turn-based averaging, leaving them susceptible to the nuances of dialogue dynamics. Meanwhile, FGA exhibits an elevated sensitivity to individual errors. GCA steps in to bridge this gap by comprehensively addressing dialogue dynamics without excessively penalizing or rewarding models.

\section{Granular Change Accuracy}
In developing GCA, our primary objective is to rectify the limitations outlined in the preceding section, as illustrated in \Cref{Fig:main_fig}. Contrary to traditional metrics that emphasize turns, GCA focuses on slots, specifically evaluating them only when their value undergoes a modification in the latest turn. This design choice tackles the issues of \textit{0/1 scores} and \textit{double counting}. Furthermore, by averaging over the total number of such modifications, GCA shifts away from \textit{turn-centric scores}. The name "Granular Change Accuracy" encapsulates its essence: a metric dedicated to evaluating accuracy based on granular changes in the belief state. A notable strength of GCA is its resilience against biases introduced by the temporal location of an error, whether it occurs early or late, or within a turn characterized by many or few active slots.

\begin{algorithm}[htb!]
	\caption{Calculating missed (M), wrong (W), overshot (O), and correct (C) predictions.} 
	\begin{algorithmic}[1]
        \small
            \State $G_{-1}$ = [],$P_{-1}$= []
            \State M, W, O, C = 0 
		\For {$t=0,1,\ldots$}
                \State Get $G_t$ and $P_t$ for turn t.
                \State $G'_t= G_t \setminus G_{t-1}$, $P'_t= P_t \setminus P_{t-1}$
                \State Cset, Wset = 0
                \For {${s,v}$ pair in $G'_t$}
                    \If {$s$ not in $P_t$}
                        \State M += 1
                    \ElsIf{$\{s,v\}$ not in $P_t$}
                        \State W += 1
                        \State add $s$ to Wset
                    \Else {}
                        \State C += 1
                        \State Add $s$ to Cset
                    \EndIf
                \EndFor

                \For {${s,v}$ pair in $P'_t$}
                    \If {$s$ not in $G_t$}
                        \State O += 1
                    \ElsIf{$\{s,v\}$ not in $P_t$ \& s not in Wset}
                        \State W += 1
                    \ElsIf{s not in Cset}
                        \State C += 1
                    \Else
                        \State continue
                    \EndIf
                \EndFor
		\EndFor
        \State return M, W, O, C
	\end{algorithmic} 
 \label{four_metrics}
\end{algorithm}

% Min: Dual-layer is confusing
% Set apart, distinguishing two separate purposes/functions

A critical distinction in GCA's design is its recognition of the two-step prediction required in DST, a nuance often overlooked in previous metrics. Instead of merely categorizing predictions as right or wrong, GCA acknowledges that DST models first determine if a slot is active within the dialogue context, and subsequently predict its value. Consequently, GCA computes state changes using four distinct counts,  as illustrated in~\Cref{Fig:4Counts} (\textit{c.f.}~\Cref{four_metrics}):

\begin{figure}[hbt!]
\centering
\includegraphics[width=0.5\textwidth]{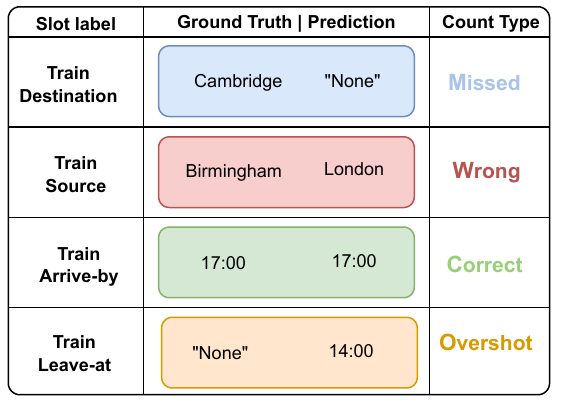}
\caption{An illustrative breakdown of the four counts used in evaluating DST model predictions. Each row represents a slot label, with corresponding ground truth and model predictions. The color-coded 'Count Type' column categorizes each prediction as 'Missed', 'Wrong', 'Correct', or 'Overshot', based on the comparison between the ground truth and the predicted values.}
\label{Fig:4Counts}
\end{figure}

Missed predictions ($M$): Number of slots valued in the ground truth Belief State (BS) but absent in the predicted BS.
Wrong predictions ($W$): Number of mismatched slot values present in both ground truth and predicted BS.
Overshot predictions ($O$): Number of slots valued in the predicted BS but absent in the ground truth BS.
Correct predictions ($C$): Number of matching slot-value pairs in both ground truth and predicted BS.
While similar counts have been outlined by \citet{smith-2014-comparative}, they presented them discretely, without consolidating them into a singular evaluation metric, a gap GCA bridges.

\begin{figure*}[htb!]
  \centering
  \small
  \begin{subfigure}[b]{0.33\linewidth}
    \includegraphics[width=\linewidth]{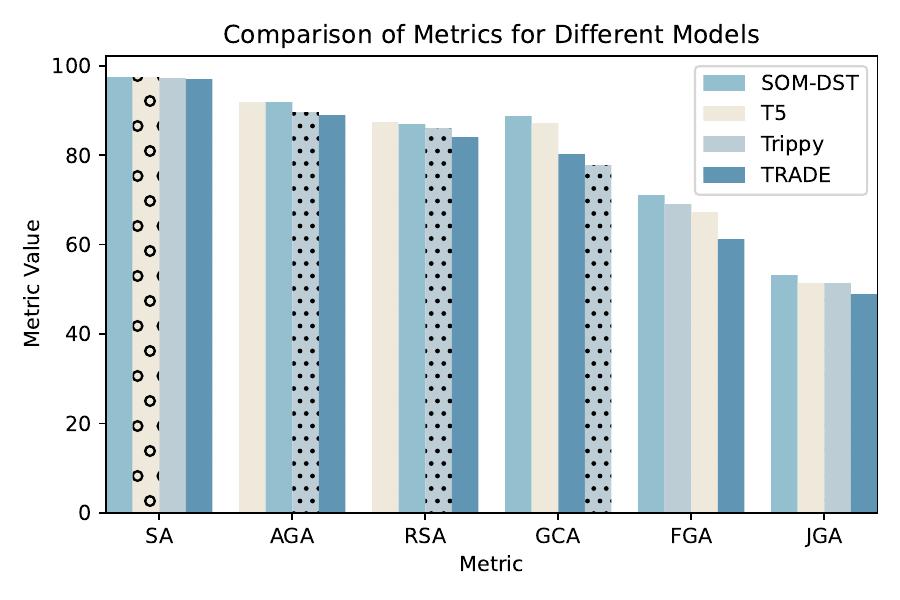}
    \caption{Full-shot results -- MultiWOZ 2.1}
    \label{Fig:full_shot_main}
  \end{subfigure}
  \hfill
  \begin{subfigure}[b]{0.33\linewidth}
    \includegraphics[width=\linewidth]{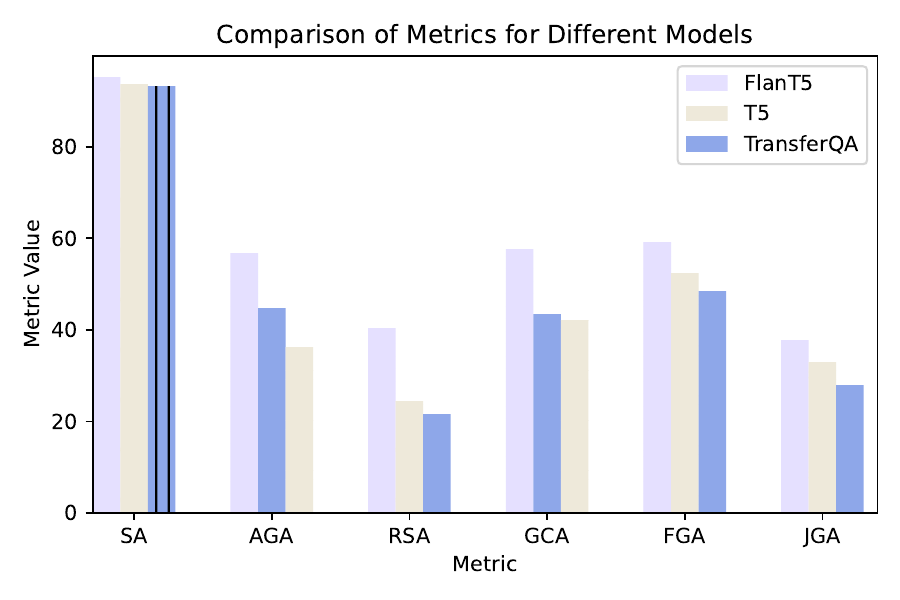}
    \caption{Zero-shot results -- MultiWOZ 2.1}
    \label{fig:zero_shot_main}
  \end{subfigure}
  \begin{subfigure}[b]{0.33\linewidth}
    \includegraphics[width=\textwidth]{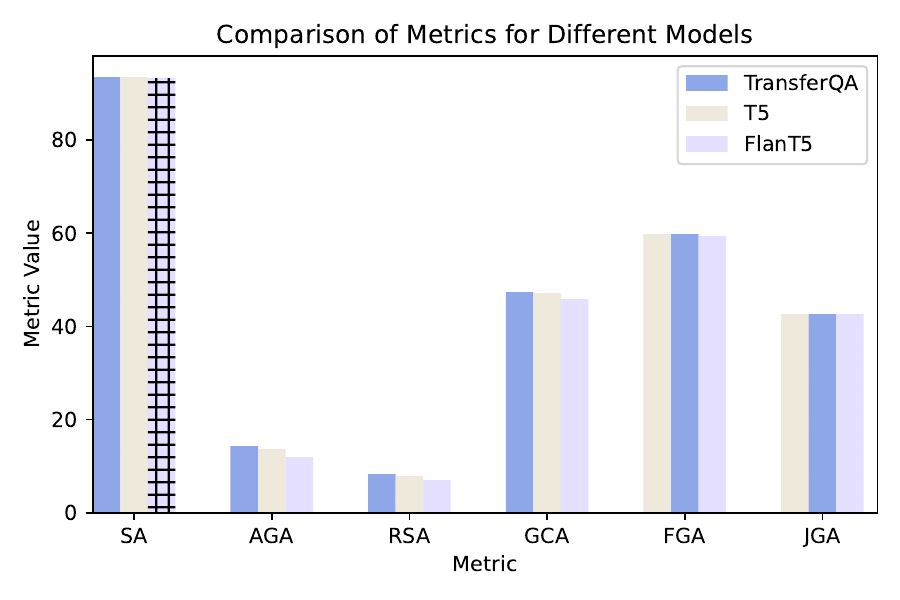}
    \caption{Zero-shot results -- SGD}
    \label{Fig:zeroshot_sgd}
\end{subfigure}
  \caption{
  Full and Zero-shot Results for MultiWOZ~2.1 and SGD datasets with various metrics.}
  \label{Fig:Main_results}
\end{figure*}

Building on the dual-layer prediction challenge inherent to DST, our proposed metric goes beyond raw counts to offer a more granular evaluation. We translate the aforementioned counts into four intermediate metrics, each tailored to assess either the act of recognizing an active slot (the slot label) or predicting its exact value (the slot value):
% Min: you need punctuation to delimit the equation from the text.

\paragraph{Value Precision: }This metric evaluates the accuracy of value predictions for detected active slots. It's formulated as: $V_P = \frac{C}{P}$, where $P=C+W+O$ represents the total number of predictions.

\paragraph{Value Recall:} This metric evaluates how well the model recalls actual slot values from the dialogue context: $V_R = \frac{C}{G}$, where $G= C + W +M$ is the total number of gold values.
\paragraph{Label Precision: }This metric captures the precision with which the model identifies a slot as active, irrespective of the value's accuracy: $L_P = \frac{C+W}{P}$
\paragraph{Label Recall: }Analogous to Label Precision, this measure gauges the model's ability to recall active slots: $L_R = \frac{C+W}{ G}$ \\

The combined counts $C+W$ pertain to instances where the slot detection was correct, even if the subsequent value prediction might not be accurate. 

These intermediate products help dissect model performance at both levels of DST prediction, offering insights not captured by previous monolithic accuracy scores.
% \noindent $C+W$ refers to predictions in which the slot was correctly predicted, but the accuracy of the value prediction may vary.
Finally,  we formulate GCA using a weighted harmonic mean of the four intermediate metrics as adopted in the F1 score for its efficacy in balancing precision and recall~\footnote{We use weighted harmonic mean to weigh value accuracies differently from label accuracies. Since value accuracy is an exact match whereas label accuracy is a partial match we believe the former should have a higher value. Thus, in our experiments, we set the weight of $V_P$ and $V_R$ to $0.9$, whereas the weight of $L_P$ and $L_R$ to $0.1$.}:
\begin{equation}
    GCA = Harmonic\_mean(V_P,V_R,L_P,L_R)
\end{equation}
% \begin{equation}
%     H(x_1, x_2, ... x_n) = \frac{n}{(\frac{1}{x_1}+ \frac{1}{x_2} + ... + \frac{1}{x_n})}
% \end{equation}

For the two predictions $P1$ and $P2$ in \Cref{Fig:main_fig}, GCA attributes significantly higher performance to the former reflecting the real performance they depict unlike all previous metrics.

% \begin{equation}
%     GCA = \frac
%         {
%             (P+G)
%         }
%         {   
%             \frac{P*\alpha}{VP} + 
%             \frac{G*\alpha}{VR} +
%             \frac{P*(1 - \alpha)}{LP} +
%             \frac{G*(1 - \alpha)}{LR}
%         }
% \end{equation}

\section{Experiments and Analysis}

% \kmy{Need to motivate why you are doing these experiments.  Are you trying to show GCA makes sense? Is robust? Has different results than prior metrics?}\taa{Edited, please check.}
To study how GCA's design affects popular benchmarks and assess its effectiveness in addressing previous metrics' under/overestimation tendencies, we conduct experiments with MultiWOZ 2.1 and SGD datasets, evaluating 6 DST models: \texttt{TRADE}~\cite{wu-etal-2021-transferable}, \texttt{SOM-DST}~\cite{kim-etal-2020-efficient}, \texttt{Trippy}~\cite{heck-etal-2020-trippy},\texttt{T5} based model by \citet{lin-etal-2021-leveraging}, \texttt{TransferQA}~\cite{lin-etal-2021-zero} and \texttt{FlanT5}~\cite{Chung2022ScalingIL}. For \texttt{TRADE} and \texttt{SOM-DST} we re-use the predictions reported in \citet{dey-etal-2022-towards}. 
% We trained Trippy, TransferQA, FlanT5, and T5 models from scratch on an NVIDIA-V100 using the best hyperparameter settings reported by the authors.

\subsection{Benchmarking Results}

% \setlength{\tabcolsep}{3pt}
% \begin{table}[tbh]
% \small
%     \centering
%     % \fontsize{8pt}{8pt}\selectfont
%     \begin{tabular}{ccccccc}
%         \hline
%         Model &   JGA & FGA & SA & AGA &  RSA & GCA \\
%         \hline
%         \textbf{Full-shot}&\multicolumn{6}{c}{\textbf{MultiWOZ 2.1}} \\
%         \hline        
%         TRADE &   48.86 & 61.19 & 96.96 & 88.79 & 83.87 & 80.15 \\
%         SOM-DST &   53.09 & 71.04 & 97.36 & 91.71 & 86.91 & 88.63  \\
%         Trippy &   51.22 & 68.87  & 97.09 & 89.47 & 86 & 77.72  \\
%         T5 & 51.4 &  67.27 & 97.32 & 91.72 & 87.25 & 87.19 \\
%         \hline
%         \textbf{Zero-shot}&\multicolumn{6}{c}{\textbf{MultiWOZ 2.1}} \\
%         \hline
%         TransferQA & 27.86 & 48.37  & 93.16 & 44.56 & 21.42 & 43.27 \\
%         T5 & 32.89 &  52.22 & 93.56 & 36.21 & 24.37 & 42.08 \\
%         FlanT5 & 37.58 & 59.00  & 95.10 & 56.76 & 40.27 & 57.63 \\
%         \hline
%         \textbf{Zero-shot}&\multicolumn{6}{c}{\textbf{SGD}} \\
%         \hline
%         TransferQA & 42.61 & 59.75 & 93.36 & 14.3  & 8.25  & 47.21 \\
%         T5 & 42.62 & 59.76 & 93.34 & 13.64 & 7.87  & 47.09 \\
%         FlanT5 & 42.45 & 59.24 & 93.24 & 11.88 & 6.86  & 45.82 \\
%         \hline
%     \end{tabular}
%     \caption{Benchmarking results.}
%     \label{tab:results}
% \end{table}

In this section, we present comprehensive benchmarking results over the listed datasets and models under varying training conditions. We categorize our results based on full-shot, zero-shot, and few-shot training scenarios. 

\subsubsection{Full-shot Results}
\Cref{Fig:full_shot_main} shows the full-shot results.
% (for numerical results, \textit{c.f.} upper half of~\Cref{tab:results}~\tatodo{Make sure setting col sep to 3 pt is legal.} \tatodo{This is verbose in combination with Fig 2}).
We set $\lambda=0.5$ for FGA following~\citet{dey-etal-2022-towards}. Notably, JGA and FGA tend to produce lower performance scores owing to their binary scoring approach. On the contrary, SA and AGA exhibit inflated scores, reflecting their tendency to overestimate. GCA and RSA scores are positioned between these two extremes. It's noteworthy to point out the fluctuations in model rankings based on different metrics. For example, while GCA ranks \texttt{SOM-DST} as the top-performing model, RSA places \texttt{T5} at the top. This emphasizes that the choice of evaluation metric can indeed lead to distinct model hierarchies.
\subsubsection{Zero-shot Results}
Zero-shot results are shown in \Cref{fig:zero_shot_main} and \Cref{Fig:zeroshot_sgd}. One can observe that various weaknesses described in \Cref{Fig:main_fig} are at work in these results.
% Min2: you may have space to add some of this back in if you want.
% Because ground-truth dialogue states in the zero-shot scenario tend to have many turns with inactive (\textit{i.e.} ``none''-valued) slots the multiple-counting score problem is observed pushing JGA to the lower end of the scores.
% The presence of numerous inactive (\textit{i.e.} ``none''-valued) slots in ground-truth dialogs of MultiWOZ 2.1 during the zero-shot scenario leads to the issue of multiple-counting scores, which ultimately results in lower scores for the JGA. 
Due to the cross-domain nature of MultiWOZ, most turns within zero-shot adaptation do not have any active slots. This leads to the issue of double-counting (early mistakes are multiply counted for most turns), which ultimately results in lower scores for JGA. FGA addresses this problem by decaying the mistakes; however, the turn-centric scoring it employs counts most empty turns as successful predictions of the model, which deceptively boosts the final score of a model. RSA has scores that float at the bottom as whenever a turn does not have any active slots it scores that turn as 0 which drags the average score of the dialogue down (turn-centric scoring). GCA takes the middle ground between these metrics since it counts each mistake once at the first encounter and calculates the performance by aggregating accuracy over the model's actions rather than turns (\textit{c.f} \Cref{Sec:detailed_results} for an extensive explanation on a running example).

\begin{figure}[htb!]
  \centering
  \small
  \hfill
  \begin{subfigure}[b]{\linewidth}
    \includegraphics[width=\linewidth]{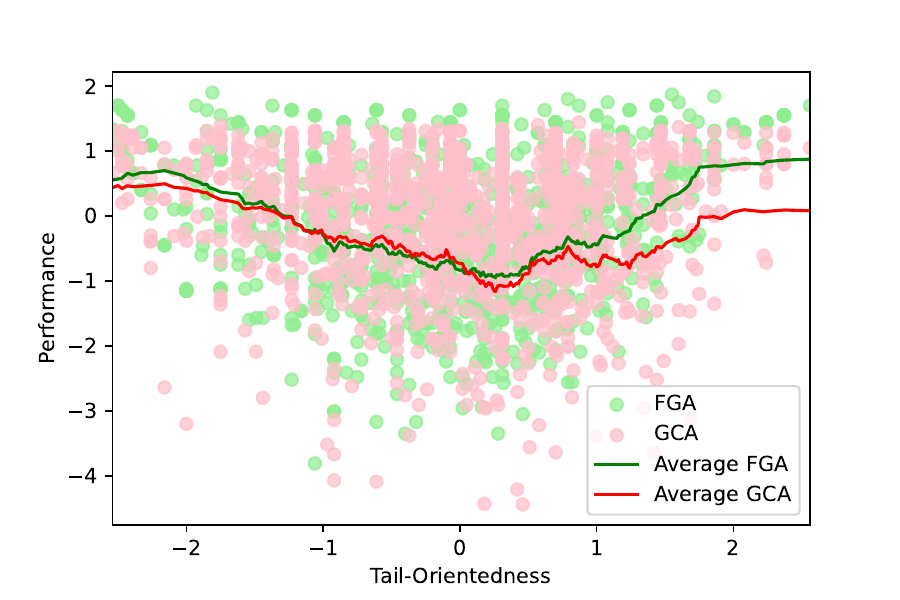}
    \caption{Performance by TO measure normalized by mean normalization.}
    \label{fig:T0}
  \end{subfigure}
  
  \begin{subfigure}[b]{\linewidth}
    \includegraphics[width=\textwidth]{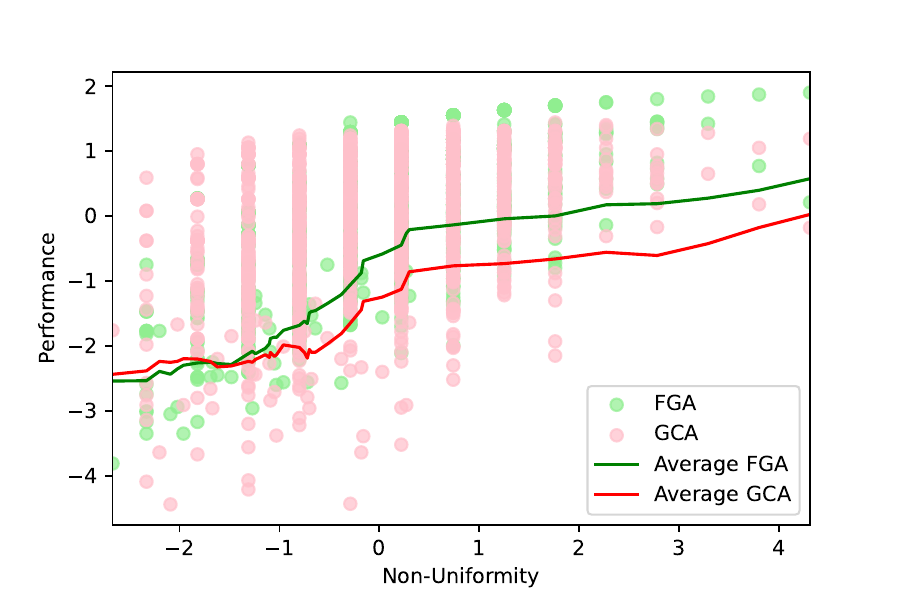}
    \caption{Performance by NU measure normalized by mean normalization.}
    \label{fig:NU}
\end{subfigure}
  \caption{
  \ref{fig:T0} and \ref{fig:NU} depict how spurious traits affect GCA and FGA scoring.}
  \label{Fig:Analysis}
\end{figure}

\begin{figure*}[hbt!]
    \centering
    \includegraphics[width=0.9\linewidth]{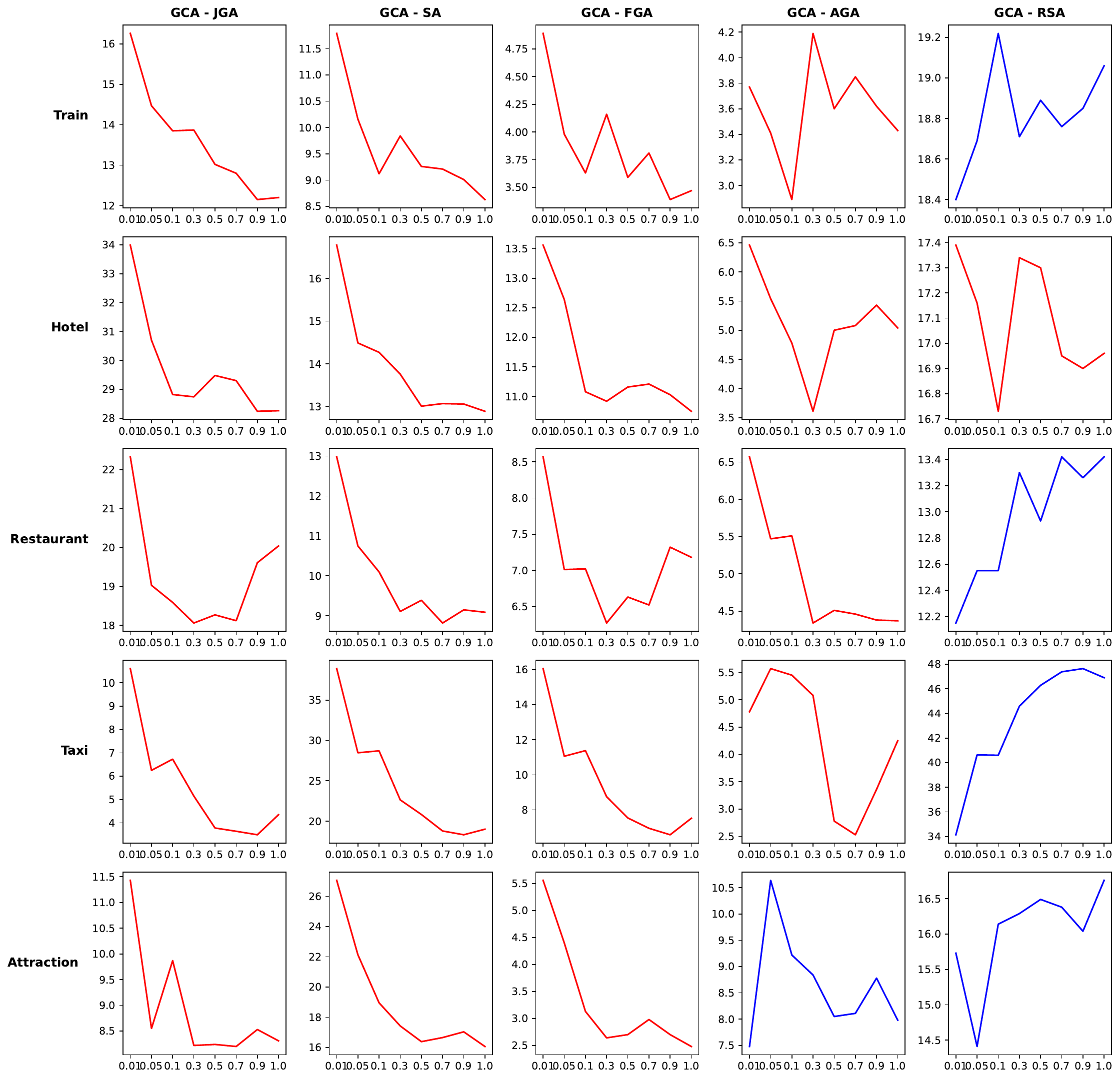}
    \caption{Variation in the absolute difference between GCA and other metrics based on the ratio of shots used during training. Red plots signify a decreasing difference, while blue plots denote an increasing difference as shots increase.}
    \label{fig:few_shot_all}
\end{figure*}

\subsubsection{Few-shot Results}
% We further explore the differences emerging between zero and full-shot for GCA versus JGA by slowly increasing the number of shots. \Cref{fig:few_shot_all} shows the results for the MultiWOZ dataset with T5 model\tatodo{Double check the model used for few-shot experiments}. We see that the difference in GCA and JGA metrics gets larger as the number of shots decreases, \textit{i.e.} as the model's error rate increases. This further supports our hypothesis that JGA's harsh penalizing s
In an effort to further understand the disparities between zero and full-shot for GCA in comparison to other metrics, we incrementally increased the number of shots used during training. The results, for the MultiWOZ dataset with the T5 model, are displayed in \Cref{fig:few_shot_all}. A clear observation is that the disparity between GCA and other metrics intensifies as the number of shots diminishes, that is, as the model's error rate climbs. From the figure, we derive the following observations:

\paragraph{Predominant Effect:} A significant majority of the metrics exhibit increased differences in comparison to GCA when models are trained with scarcer data. In the figure, this trend is depicted through color: red plots indicate a higher difference, while blue plots signify a lower difference. This reinforces the idea that for the majority of metrics, the identified weaknesses become more pronounced when evaluating models trained on limited data.

\begin{figure}[hbt!]
    \centering
    \includegraphics{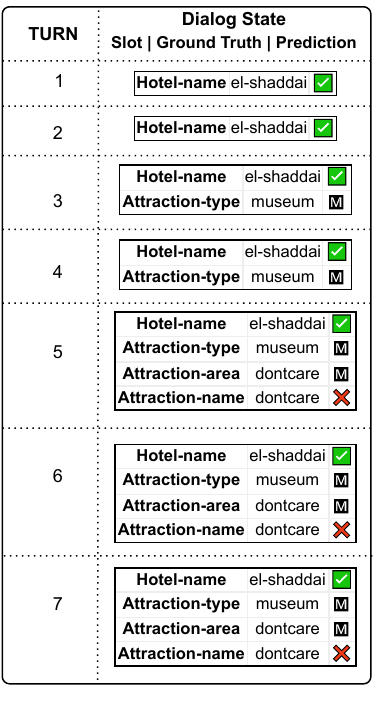}
    \caption{Sample dialogue from MultiWOZ 2.1 dataset, (MUL1110) with ground truth and predicted belief states. GCA: 31.43, FGA: 48.84}
    \label{fig:analysis_1}
\end{figure}

\begin{figure}[hbt!]
    \centering
    \includegraphics{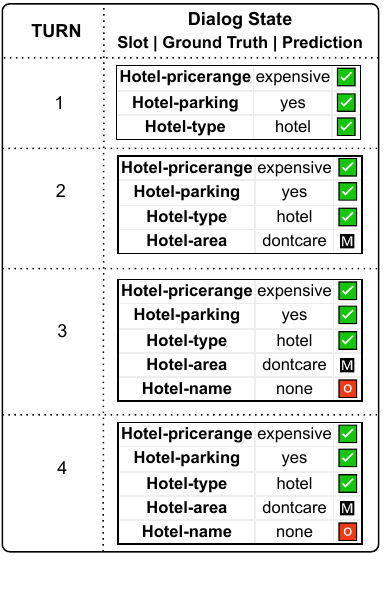}
    \caption{Sample dialogue from MultiWOZ 2.1 dataset, (SNG0779) with ground truth and predicted belief states. GCA: 75, FGA: 34.84}
    \label{fig:analysis_2}
\end{figure}
% \paragraph{Comparison with FGA \& JGA:} For zero-shot scenarios, the performance disparity between GCA and FGA/JGA narrows as the shot count ascends, irrespective of the domain. 
% \tatodo{TBW}
% % Note that in general FGA shows higher and JGA shows lower scores in zero-shot compared to GCA. However the consistent change in absolute differences with both metrics point to increasing over and underestimation tendencies of both metrics.
% \paragraph{Comparison with SA \& AGA:} Both SA and AGA register higher scores compared to GCA. However, it's worth noting that the differences between them and GCA shrink with an increase in training shots.
\paragraph{RSA's Unique Behavior:} Contrary to other metrics, RSA showcases an inverse effect. Notably, it produces markedly lower scores in the zero-shot scenarios (as seen in 2-b and 2-c). This is largely due to its behavior of assigning a score of zero for turns without active slots—a frequent occurrence in zero-shot evaluations—subsequently dragging the overall dialogue score downward.

In essence, the variance in metrics underscores the need for nuanced evaluations, especially in few-shot scenarios, to garner insights into a model's actual capabilities and limitations.

\subsection{Fine-Grained Analysis}
\label{spurTraits}

To analyze edge cases, we examined 20 predictions of \texttt{TRADE} and \texttt{SOM-DST} models where FGA and GCA show the largest disagreement. 
We chose FGA for detailed examination because of its one of the most recent metrics, contrasted to community standard JGA (Joint Goal Accuracy).  We observe that FGA overestimates the performance when errors are accumulated in a few turns, \textit{i.e.} the mistakes are \textbf{not uniformly} distributed. Especially if these accumulations occur in the later part of the dialog, \textit{i.e.} when the mistakes show a \textbf{tail-oriented} distribution (c.f. samples in~\Cref{sample_dialogues}).

\paragraph{Tail-Oriented Mistake Distribution.}
To examine the impact of tail-oriented mistakes on FGA and GCA evaluation, we introduce a new measure, $TO$:
\\
\begin{equation}
    TO = \frac{E_t - (\frac{n-1}{2})}{n}
\end{equation}
\begin{equation}
    E_t = \frac{\sum_{i=0}^m {t_i}}{m}
\end{equation}
, where $n$ is the total number of turns, $t_i$ is the turn index of mistake $i$, and $m$ is the total number of mistakes.
It calculates the average distance of each mistake's turn from the middle turn of the dialog. \Cref{fig:T0} illustrates the performance distribution based on $TO$. Despite GCA consistently yielding higher results for lower values of $TO\approxeq(-2.5)$ to $(-1.0)$, we observe that FGA shows similar or even higher scores at the right-hand side of the figure. This suggests that as dialog state mistakes become more tail-oriented, FGA tends to overestimate the performance.

\paragraph{Non-Uniform Mistake Distribution.} 
In a similar manner we define a non-uniformity measure, NU, 
% $NU = \frac{\sum_i^T |m_i - E|}{E}$
\begin{equation}
NU = \frac{\sum_{t=0}^n |m_t - E_m|}{E_m}
\end{equation}
\begin{equation}
E_m = \frac{m}{n}
\end{equation}
, where $m_t$ is the number of mistaken predictions (missed, over-shot or wrong) in turn $t$. $E_m$ is the expected number of mistakes per turn under a uniform distribution, and $n$ is the total number of turns. Figure~\ref{fig:NU} demonstrates the performance distribution by $NU$. The results are mean-normalized, causing the NU measure to have otherwise impossible negative values. For lower $NU\approxeq(-2.5$) to $(-1.5)$, FGA generally exhibits lower values compared to GCA, however, one can observe it going higher as $NU$ values increase --- \textit{i.e.} for $NU \geq (-1.0)$. This suggests that FGA is adversely affected by the uniform spread of prediction errors.

Finally, we further calculate the Pearson Correlation Coefficients of both FGA and GCA with both spurious traits across dialogs. The correlations between $TO$ and FGA/GCA are $0.08$/$-0.05$, whereas between $NU$ and FGA/GCA are $0.59$/$0.40$ respectively. The differences between these correlations are significant according to \citet{zous-conf-intervals}'s confidence interval tests. FGA's correlation with both features is significantly stronger with a 95\% confidence level. These results show that GCA is less susceptible to spurious features than FGA.

\subsection{Sample GCA and FGA Scores}
\label{sample_dialogues}

This section presents two sample dialogs from the MultiWOZ 2.1 dataset along with DST model predictions, and FGA/GCA evaluations.

The dialogue in ~\Cref{fig:analysis_1} is an example where FGA overestimates the performance of a dialogue even though it only predicts one out of four slots correctly. This is because the majority of mistakes in the dialogue occur closer to the tail of the dialogue and correct prediction done at the first turn is counted multiple times --- \textit{i.e.} \textit{Double-counting score} weakness.

~\Cref{fig:analysis_2} on the other hand presents an example where FGA underestimates the performance of a model even though it predicts three out of five slots correctly -- significantly better compared to the performance in~\Cref{fig:analysis_1}. This is because, unlike the prior example, the dialog is shorter, leading to fewer repeated predictions by FGA.

\subsection{Fine-grained Explanation of Zero-shot Results}
\label{Sec:detailed_results}
\begin{figure}[hbt!]
    \centering
    \includegraphics{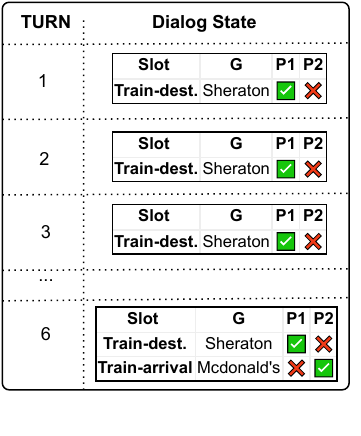}
    \caption{Hypothetical dialog}
    \label{fig:analysis_3}
\end{figure}
In zero-shot evaluations, given their cross-domain nature focused on a single domain, most turns don't contain active slots. This characteristic poses a challenge for existing metrics, leading to skewed evaluations. To elucidate, let's delve into a hypothetical dialogue depicted in~\Cref{fig:analysis_3}. This dialogue has two active slots: one at the first turn and another at the last. Such scenarios aren't uncommon in datasets like MultiWOZ, where, for instance, taxi domain dialogues primarily discuss hotel or restaurant bookings, leaving taxi booking for the final turns.

Now, let's analyze two prediction scenarios for this dialogue, represented as $P1$ and $P2$. In $P1$, the first slot is predicted correctly, but the final slot is amiss. Conversely, $P2$ nails the final slot prediction but misses the initial one. With JGA, scores for $P1$ and $P2$ are $83.33$ and $0$, respectively. Such an overestimation for $P1$ arises because JGA rewards every empty turn, while $P2$ is harshly penalized due to JGA's all-or-nothing approach. RSA paints a similar picture, scoring $91.67$ for $P1$ and $8.33$ for $P2$. FGA nuances it a bit with $83.33$ for $P1$ and $59.75$ for $P2$. Though it aligns with JGA for $P1$, it is more generous for $P2$ by positively scoring empty turns, explaining FGA's inflation in zero-shot contexts. Contrarily, GCA offers a balanced perspective with $52.38$ for both, as it gauges changes in dialogue state uniformly, penalizing and rewarding predictions only once.

% \subsection{Discussion of Validity of Weaknesses}

% \taa{Regarding the idea that double-counting scores should be a trait of performance metrics:}\tatodo{Thought here are in very raw form, clear them and make the writing fluent.}If the DST training and evaluation followed a reinforcement learning setting where the mistakes that the model made had an instant effect on the generated response this concern would be viable. However, both the Multiwoz and SGD benchmarks rely upon the assumption that the DST model works perfectly and thus the generated response has access to an ideal DST model. Given this assumption there is no way that DST models that we train can get feedback of an earlier mistake they did within the dialogue context. Thus I believe imposing such a penalty on the models predictions is somewhat superficial.

\section{Conclusion}
In this work, we have delved deep into the inherent weaknesses of prevalent DST evaluation metrics. Specifically, we spotlighted their propensity to over or underestimate model performance, the pitfalls of 0/1 scoring, the turn-centric nature of their scoring systems, and the double-counting of errors. Addressing these shortcomings, we introduce GCA, a novel metric that prioritizes accuracy based on belief state changes, offering a more nuanced evaluation approach than simple turn-by-turn assessments. Through rigorous analyses, we demonstrate that GCA provides a more balanced and representative evaluation, effectively sidestepping the pitfalls within other metrics. Additionally, GCA showcases a notably diminished correlation with certain dialog traits that shouldn't influence metric performance, such as the non-uniform distribution or tail-skewness of mistakes. Most crucially, our results emphasize GCA's robustness in evaluations of models trained under data constraints. In the complex landscapes of few-shot and zero-shot learning, where standard measurements become unstable due to increased model errors, GCA stands out as a reliable benchmark for precise assessment. We are convinced that the DST community will see long-term benefits from integrating GCA into their array of metrics for model benchmarking.

\section{Limitations}
We list two main limitations of our work as follows:

\noindent\textbf{$1$) We do not address partial credits at the slot level.} Though GCA is more exhaustive than existing metrics, there is still room for improvement by partial credit of slot values; \textit{i.e.} by calculating the similarity of ground-truth and predicted values.

\noindent\textbf{$2$) We do not report results with larger models.}  We have not conducted experiments with larger models. It would be interesting to see whether these results are consistent among different sizes of models.

\section{Acknowledgements}
This research was supported by the SINGA scholarship from A*STAR. We wish to express our profound gratitude to Prof. Min-Yen Kan from the National University of Singapore, School of Computing, for his invaluable contributions. This work would not have been possible without his support at both the ideation and writing stages. We would also like to thank anonymous reviewers for their insightful feedback on how to improve the paper.

\section{Bibliographical References}\label{sec:reference}

\bibliographystyle{lrec-coling2024-natbib}
% \bibliography{custom, anthology}
\bibliography{custom}

% \section{Language Resource References}
% \label{lr:ref}
% \bibliographystylelanguageresource{lrec-coling2024-natbib}
% \bibliographylanguageresource{languageresource}

\end{document}